\documentclass{article}

\usepackage[final,main]{neurips_2026}

\usepackage[utf8]{inputenc}
\usepackage[T1]{fontenc}
\usepackage[hypertexnames=false]{hyperref}
\hypersetup{hidelinks}
\usepackage{url}
\usepackage{booktabs}
\usepackage{amsfonts}
\usepackage{amsmath,amssymb}
\usepackage{nicefrac}
\usepackage{microtype}
\usepackage{xcolor}
\usepackage{graphicx}
\usepackage{pgfplots}
\usepackage{pgfplotstable}
\pgfplotsset{compat=1.17}
\usepackage{enumitem}
\usepackage{algorithm}
\usepackage{algorithmic}
\usepackage{multirow}
\usepackage{colortbl}
\usepackage{tikz}
\usetikzlibrary{positioning,arrows.meta,calc,patterns,decorations.pathreplacing}

\definecolor{topkBlue}{HTML}{2563EB}
\definecolor{sosRed}{HTML}{DC2626}
\definecolor{contrastOrange}{HTML}{EA580C}
\definecolor{entropyPurple}{HTML}{7C3AED}
\definecolor{varianceGreen}{HTML}{16A34A}
\definecolor{geluTeal}{HTML}{0D9488}
\definecolor{swishAmber}{HTML}{D97706}
\definecolor{reluGray}{HTML}{6B7280}
\definecolor{lightgray}{HTML}{F3F4F6}
\definecolor{bestcell}{HTML}{DBEAFE}
\definecolor{burstTeal}{HTML}{0F766E}

\title{Selectivity and Shape in the Design of Forward-Forward Goodness Functions}

\author{%
  \begin{tabular}{cc}
    Talha Rüzgar Akkuş\thanks{Equal contribution.} & Şuayp Talha Kocabay\footnotemark[1] \\
    Poindexter Labs & Poindexter Labs \\
    \texttt{tra@poindexterlabs.ai} & \texttt{stk@poindexterlabs.ai} \\[0.2cm]
    Kamer Ali Yüksel\footnotemark[1] & Hassan Sawaf \\
    Feedzai & aiXplain, Inc. \\
    \texttt{kamer.yuksel@feedzai.com} & \texttt{hassan@aixplain.com}
  \end{tabular}
}

\begin{document}

\maketitle

\begin{abstract}
The Forward-Forward (FF) algorithm trains networks layer-by-layer using a local
``goodness function,'' yet the choice of activity statistic remains an important
design question. We systematically study how selectivity and activity shape affect
FF learning, introducing selective goodness functions based on top-$k$ and
entmax-weighted energy alongside shape-sensitive objectives based on normalized
higher-order moments. In particular, we propose \emph{burstiness goodness}, a
parameter-free, scale-invariant excess-kurtosis objective that rewards heavy-tailed
activity patterns. Our study examines goodness functions together with activation
and label-injection choices across six small-image datasets, and uses controlled
experiments to isolate their effects. In validation-tuned, five-seed Fashion-MNIST
experiments with matched ReLU activation, standard label injection, and per-layer
scoring, burstiness improves accuracy from $73.2\pm0.9\%$ to $88.6\pm0.4\%$
over sum-of-squares. On CIFAR-10, burstiness improves block-local ResNet-18
accuracy from $28.9\pm0.5\%$ to $46.8\pm2.5\%$ under the same separate
per-block label pathway. These results show that activity-shape statistics can
provide effective local learning signals across the tested settings.
Code is available at \url{https://github.com/suayptalha/ff-selectivity-shape}.

\end{abstract}

\section{Introduction}
\label{sec:intro}

The Forward-Forward (FF) algorithm~\citep{hinton2022forward} replaces backpropagation's
global backward pass with a local, layer-wise learning rule: each layer maximizes a
scalar ``goodness'' for positive (correctly labeled) data and minimizes it for negative data.
\citet{hinton2022forward} defined goodness as the sum of squared activities (SoS),
and this energy-based choice has been widely used
\citep{10191727,lorberbom2024layer,lee2023symba}.
A recent benchmark~\citep{shah2025goodness} evaluated 21~goodness functions but within a
fixed architecture; no prior work has jointly studied goodness functions, activation
functions, label-injection strategies, and the underlying principle governing what makes
a good goodness function.

We argue this gap is significant: the goodness function defines each layer's objective
landscape, determining what representations are rewarded and what features emerge.
Sum-of-squares aggregates all neurons and measures total activity magnitude.
In contrast, we show that shape-sensitive metrics can better capture the nuanced,
non-Gaussian signatures of discriminative features.
We conduct a systematic study of the following \textbf{design hypothesis}:
sensitivity to the shape of neural activity can improve FF goodness objectives
within the tested settings. Selective functions (top-$k$, entmax-weighted energy)
retain magnitude information while focusing on peak activity; burstiness and
normalized higher moments instead measure scale-invariant distributional shape.
In matched standard-FF experiments, burstiness improves Fashion-MNIST accuracy
by 15.4 percentage points over validation-tuned SoS with ReLU held fixed.
Our contribution is the formulation and controlled study of these established
statistics as local FF objectives, rather than the invention of the statistics themselves.

\paragraph{Contributions.}
\begin{enumerate}[leftmargin=*,topsep=2pt,itemsep=1pt]
\item We formulate selective top-$k$ and entmax-weighted energy objectives and
scale-invariant burstiness goodness (\S\ref{sec:method}).
\item We isolate objective effects through validation-tuned, five-seed comparisons
with matched activation, label pathway, and standard inference scoring
(\S\ref{sec:main_results}).
\item We study activation and label-pathway interactions and three parameter sweeps,
and test transfer to convolutional and residual local blocks
(\S\ref{sec:sparsity_sweep}--\ref{sec:spatial}).

\end{enumerate}

\section{Related Work}
\label{sec:related}

\paragraph{Forward-Forward learning.}
\citet{hinton2022forward} introduced FF; subsequent work studied layer sizes and
negative data~\citep{10191727}, layer collaboration~\citep{lorberbom2024layer},
symmetric contrastive variants~\citep{lee2023symba}, predictive
coding~\citep{ororbia2023predictive}, and per-layer label injection
(FFCL;~\citealt{srinivasan2024forward}).
\citet{shah2025goodness} benchmarked 21~goodness functions within a fixed architecture
with peer normalization and downstream classifiers.
Our work differs by jointly studying the goodness function, activation function,
label pathway, and the scoped shape-sensitivity hypothesis.

\paragraph{Complete methods versus objective controls.}
SCFF~\citep{chen2025scff} changes positive/negative construction and representation
learning; Mono-Forward~\citep{gong2025mono} uses local classification losses.
SymBa and FFCL likewise modify more than the activity statistic.
Our loss substitutions do not benchmark these complete pipelines. Convolutional and
scalable FF work~\citep{scodellaro2023conv,krutsylo2025scalable,gong2025asge}
addresses broader architecture and scaling questions; our spatial tests establish
relative objective gains in matched local blocks, not superiority to those methods.

\paragraph{Sparse transformations.}
The $\alpha$-entmax family~\citep{martins2016softmax,correia2019adaptively,peters2019sparse}
produces sparse probability distributions and has been applied to attention mechanisms.
We apply entmax in a novel context: as a sparse weighting mechanism within a goodness
function, creating an adaptive alternative to hard top-$k$ selection.

\paragraph{Sparse coding and ICA.}
Sparse coding~\citep{olshausen1996emergence}, $k$-WTA
networks~\citep{ahmad2019dense,maass2000computational}, and
ICA~\citep{hyvarinen2000independent} motivate our approach.
Top-$k$ encourages kWTA-like sparsity; entmax provides a differentiable relaxation.
Burstiness borrows a non-Gaussianity statistic from ICA without claiming to recover
independent components.
The heavy-tailed distributions of deep network
activations~\citep{martin2019traditional} further motivate kurtosis as a natural
training signal.
Local learning rules (Hebbian~\citep{hebb1949organization}, contrastive
Hebbian~\citep{xie2003equivalence}, equilibrium
propagation~\citep{scellier2017equilibrium}) are orthogonal---we improve the
objective within the FF framework.

\section{Background: The Forward-Forward Algorithm}
\label{sec:background}

\paragraph{Training.}
Given input $\mathbf{x}$ and label $y$, FF creates positive/negative inputs by embedding
the correct/incorrect label:
\begin{equation}
\mathbf{x}^+ = \text{norm}([\mathbf{x};\; s \cdot \text{onehot}(y)]), \qquad
\mathbf{x}^- = \text{norm}([\mathbf{x};\; s \cdot \text{onehot}(\tilde{y})]),
\quad \tilde{y} \neq y.
\label{eq:input}
\end{equation}
Each layer $\ell$ computes $\mathbf{h}_\ell = f_\ell(\mathbf{h}_{\ell-1};\theta_\ell)$
and is trained via:
\begin{equation}
\mathcal{L}_\ell =
\mathbb{E}\!\left[\log\!\left(1 + e^{\tau - g(\mathbf{h}_\ell^+)}\right)\right]
+ \mathbb{E}\!\left[\log\!\left(1 + e^{g(\mathbf{h}_\ell^-) - \tau}\right)\right],
\label{eq:loss}
\end{equation}
where $g(\cdot)$ is the goodness function and $\tau$ a threshold.
Here $\tilde{y}$ is a randomly sampled incorrect class label.
Layers are trained independently; outputs are L2-normalized before propagation.

\paragraph{Inference.}
For each candidate label $c$, we embed and forward through all layers,
predicting $\hat{y} = \arg\max_c \sum_\ell g(\mathbf{h}_\ell^{(c)})$.

\paragraph{The goodness function.}
\citet{hinton2022forward} defined $g_{\text{SoS}}(\mathbf{h}) = \sum_{i=1}^{d} h_i^2$,
the energy statistic used in the original algorithm.
We challenge the implicit assumption that total squared activity sufficiently
summarizes a layer's representation.

\section{Goodness Design: Questions and Objectives}
\label{sec:method}

We organize the design space around three experimental questions: does selecting
peak activity help (\S\ref{sec:topk}--\ref{sec:entmax}), can scale-invariant shape
statistics improve on energy (\S\ref{sec:burstiness}), and do these effects survive
matched activations, label pathways, and architectures (\S\ref{sec:experiments})?
The matched ReLU comparison isolates a 15.4-point gain; the spatial tests examine
whether that relative objective advantage transfers beyond flattened MLPs.
FFCL and activation choices below are existing components used for interaction ablations.

\subsection{Top-$k$ Goodness}
\label{sec:topk}

\emph{Top-$k$ goodness} measures only the $k$ most active neurons:
\begin{equation}
g_{\text{top-}k}(\mathbf{h}) = \frac{1}{k}\sum_{i \in \mathcal{S}_k(\mathbf{h})} h_i,
\qquad \mathcal{S}_k(\mathbf{h}) = \text{argtop-}k(\mathbf{h}),
\label{eq:topk}
\end{equation}
with $k = \max(5, \lfloor 0.02\, d \rfloor)$ (2\% of layer width).
Unlike SoS, top-$k$ ignores the $(d{-}k)$ least active neurons, creating a focused
learning signal that encourages sparse, discriminative representations.
This design is motivated by $k$-winners-take-all (k-WTA) mechanisms
observed in biological circuits, where competitive inhibition ensures that
only the most relevant neurons respond to a given stimulus~\citep{maass2000computational}.
By optimizing for peak activity, the layer learns to allocate its 
capacity to only the most informative features (\S\ref{sec:analysis}).

\subsection{Entmax-Weighted Energy Goodness}
\label{sec:entmax}

Whereas top-$k$ applies hard selection, \emph{entmax-weighted energy} uses
$\alpha$-entmax~\citep{correia2019adaptively,peters2019sparse} for adaptive sparse
weighting:
\begin{equation}
g_{\text{entmax}}(\mathbf{h};\, \alpha) = \sum_{i=1}^{d} \pi_i\, h_i^2,
\qquad \boldsymbol{\pi} = \text{entmax}_\alpha(\mathbf{h}).
\label{eq:entmax}
\end{equation}
The parameter $\alpha$ controls sparsity ($\alpha{=}1$: softmax/dense;
$\alpha{=}2$: sparsemax~\citep{martins2016softmax}/hard sparse), and entmax
\emph{learns} how many neurons are relevant per input.

\subsection{Burstiness (Excess Kurtosis) Goodness}
\label{sec:burstiness}

Orthogonally to selective measurement, \emph{burstiness goodness} measures a
scale-invariant statistic of the full activation distribution---the excess kurtosis:
\begin{equation}
g_{\text{burst}}(\mathbf{h}) =
\frac{\frac{1}{d}\sum_{i=1}^{d}(h_i - \mu)^4}
     {\left[\frac{1}{d}\sum_{i=1}^{d}(h_i - \mu)^2\right]^2} - 3,
\label{eq:burstiness}
\end{equation}
where $\mu = \frac{1}{d}\sum_i h_i$, for nonzero variance.
Applying LayerNorm before this statistic is mathematically redundant (up to
numerical stabilization); we do not treat LN-burstiness as a separate objective.
The key property is \emph{scale invariance}: $g_{\text{burst}}(\alpha \mathbf{h}) =
g_{\text{burst}}(\mathbf{h})$ for any $\alpha > 0$, making the ideal statistic invariant to
uniform magnitude changes; numerical stabilization can break exact invariance.
It rewards heavy-tailed (``bursty'') activity patterns
analogous to cortical burst firing~\citep{lisman1997bursts}, connecting
directly to ICA where kurtosis maximization extracts independent
features~\citep{hyvarinen2000independent}.
We generalize to the $p$-th central moment:
\begin{equation}
g_{\text{moment-}p}(\mathbf{h}) =
\frac{\frac{1}{d}\sum_{i=1}^{d}(h_i - \mu)^p}
     {\left[\frac{1}{d}\sum_{i=1}^{d}(h_i - \mu)^2\right]^{p/2}} - \beta_p,
\label{eq:moment}
\end{equation}
where $\beta_p = (p{-}1)!!$ for even $p$ and $0$ for odd $p$.
At $p=4$ this recovers burstiness; higher $p$ amplifies extreme activations.

\subsection{Additional Goodness Functions}
\label{sec:additional_goodness}

We also evaluate:
\emph{contrast top-$k$}
($g = \frac{1}{k}\sum_{\mathcal{S}_k^+} h_i - \frac{1}{k}\sum_{\mathcal{S}_k^-} h_i$),
\emph{LayerNorm-top-$k$}
($g = g_{\text{top-}k}(\text{LN}(\mathbf{h}))$),
\emph{variance} and \emph{negative entropy},
and two external baselines from \citet{shah2025goodness}:
\emph{softmax-energy-margin} and \emph{game-theoretic}.

\subsection{Separate Label--Feature Forwarding (FFCL)}
\label{sec:ffcl}

In standard FF, labels are concatenated at the input only.
FFCL~\citep{srinivasan2024forward} injects class hypotheses at \emph{every} layer
via a separate projection:
\begin{equation}
\mathbf{h}_\ell = \sigma(W_\ell^{\text{feat}}\, \mathbf{h}_{\ell-1}), \qquad
\tilde{\mathbf{h}}_\ell = \mathbf{h}_\ell + W_\ell^{\text{label}}\, \mathbf{y}_{\text{oh}},
\label{eq:ffcl}
\end{equation}
where $W_\ell^{\text{label}} \in \mathbb{R}^{d \times C}$.
Goodness is computed on $\tilde{\mathbf{h}}_\ell$; only the label-free
$\mathbf{h}_\ell$ (L2-normalized) propagates to the next layer.

\subsection{Activation Functions}
\label{sec:activations}

We study five activations:
ReLU (sparse, many zeros),
GELU~\citep{hendrycks2016gelu} and Swish~\citep{ramachandran2017searching}
(smooth, dense activity),
and LN-GELU / LN-Swish (pre-activation LayerNorm + smooth nonlinearity).
Smooth activations help shape-sensitive goodness functions by providing richer
distributions, while pre-activation normalization stabilizes inputs across layers.
Algorithm~\ref{alg:ff} summarizes training.

\section{Experiments}
\label{sec:experiments}

\subsection{Experimental Setup}
\label{sec:setup}

\paragraph{Datasets.}
Primary evaluation: \textbf{MNIST}~\citep{lecun1998mnist} and
\textbf{Fashion-MNIST}~\citep{xiao2017fashionmnist} (both 10-class,
$28{\times}28$ grayscale; full combinatorial ablation).
Cross-dataset generalization:
\textbf{CIFAR-10}~\citep{krizhevsky2009learning} (3072-dim color),
\textbf{USPS}~\citep{hull1994database} (256-dim grayscale),
\textbf{SVHN}~\citep{netzer2011reading} (3072-dim color),
\textbf{EMNIST-Letters}~\citep{cohen2017emnist} (784-dim, 26 classes).
Pixel values are normalized to zero mean and unit variance.

\begin{algorithm}[H]
\caption{Forward-Forward Training with Shape-Sensitive Goodness}
\label{alg:ff}
\small
\begin{algorithmic}[1]
\REQUIRE Dataset $\mathcal{D}$, goodness function $g$, threshold $\tau$, label pathway (Std/FFCL)
\FOR{each layer $\ell = 1, \ldots, L$}
  \FOR{each mini-batch $(\mathbf{x}, y) \sim \mathcal{D}$}
    \STATE Sample wrong label $\tilde{y} \neq y$
    \IF{Standard pathway}
      \STATE $\mathbf{h}_\ell^+ \gets \sigma(W_\ell\, \mathbf{h}_{\ell-1}^+)$,
             \quad $\mathbf{h}_\ell^- \gets \sigma(W_\ell\, \mathbf{h}_{\ell-1}^-)$
             \hfill {\color{gray}// $\mathbf{h}_0^\pm$: input with label $y$ / $\tilde{y}$}
    \ELSIF{FFCL pathway}
      \STATE $\mathbf{h}_\ell \gets \sigma(W_\ell^{\text{feat}}\, \mathbf{h}_{\ell-1})$
             \hfill {\color{gray}// label-free features}
      \STATE $\mathbf{h}_\ell^+ \gets \mathbf{h}_\ell + W_\ell^{\text{label}}\, \text{onehot}(y)$,
             \quad $\mathbf{h}_\ell^- \gets \mathbf{h}_\ell + W_\ell^{\text{label}}\, \text{onehot}(\tilde{y})$
    \ENDIF
    \STATE $\mathcal{L}_\ell \gets \log(1 + e^{\tau - g(\mathbf{h}_\ell^+)}) + \log(1 + e^{g(\mathbf{h}_\ell^-) - \tau})$
           \hfill {\color{gray}// $g$: top-$k$, entmax, etc.}
    \STATE Update $\theta_\ell$ via Adam on $\mathcal{L}_\ell$
  \ENDFOR
  \STATE $\mathbf{h}_\ell \gets \text{L2-normalize}(\mathbf{h}_\ell)$
         \hfill {\color{gray}// propagate to next layer}
\ENDFOR
\end{algorithmic}
\end{algorithm}

\paragraph{Architecture \& training.}
4-layer, 2000-unit FC network (4$\times$2000, $\sim$14M params).
Adam~\citep{kingma2015adam} ($lr{=}10^{-3}$), batch 500, $\tau{=}2.0$, 60 epochs.
Negative examples use random wrong labels.
Activations are L2-normalized between layers.
The combinatorial ablation uses seed~42; matched comparisons use five seeds
(\S\ref{sec:main_results} and \S\ref{sec:spatial}).
For numerical stability, SoS is scaled by $1/d$ (mean squared activation).

\paragraph{Evaluation and protocol distinction.}
The primary controlled comparisons use only the standard sum of per-layer goodness
(\S\ref{sec:background}). Candidate-label inference requires one forward pass per class.
The exploratory grid uses the sum of per-layer goodness and goodness of concatenated
layer activations; its results are labeled accordingly. The fixed hyperparameters above
describe that grid; the matched comparisons below use validation-selected settings. Std denotes input-only label concatenation;
FFCL denotes the separate per-layer label pathway in Eq.~\ref{eq:ffcl}.

\paragraph{Experimental grid.}
We evaluate 13~goodness functions
crossed with 2~activations (GELU, Swish),
2~norm-gate settings%
\footnote{Norm-gating scales activations by $\sigma(\|\mathbf{h}\|) \cdot \mathbf{h}$.
Across all experiments, the max accuracy difference between on/off is ${<}0.4$pp;
we report the best of each pair.},
and 2~label pathways (standard, FFCL),
plus a ReLU+SoS baseline.
Additionally, we conduct three continuous sweeps: top-$k$ cardinality~$k$,
entmax parameter~$\alpha$, and moment order~$p$ (\S\ref{sec:sparsity_sweep}).

\subsection{Main Results}
\label{sec:main_results}

\textbf{Does burstiness help when other design choices are held fixed?}
We use a held-out 55k/5k/10k train/validation/test split on Fashion-MNIST.
Each objective/activation/pathway combination receives the same $3\times3$
validation search over label scale and threshold, followed by five test seeds.
Table~\ref{tab:matched} holds architecture, activation, label pathway, and scoring
fixed within each row. Standard-FF ReLU isolates a 15.4-point improvement;
the per-layer-label GELU comparison isolates 5.8 points. Thus the gain survives
stronger SoS tuning and does not require FFCL or concatenated-activation scoring.

\begin{table}[H]
\centering
\caption{Matched Fashion-MNIST accuracy (\%, mean $\pm$ standard deviation,
five seeds; 4$\times$2000; standard per-layer scoring). Each objective receives
the same validation-search budget. Gains use unrounded means.}
\label{tab:matched}
\small
\begin{tabular}{llccc}
\toprule
Pathway & Activation & Tuned SoS & Burstiness & Gain (pp) \\
\midrule
Std & ReLU & $73.2\pm0.9$ & $88.6\pm0.4$ & +15.4 \\
Std & GELU & $61.5\pm0.4$ & $87.9\pm0.4$ & +26.3 \\
FFCL & GELU & $82.1\pm0.1$ & $87.9\pm0.2$ & +5.8 \\
\bottomrule
\end{tabular}
\end{table}

\paragraph{Scoring ablation.}
For burstiness, standard versus concatenated-ensemble scoring gives
$88.6\pm0.4$ versus $88.2\pm0.4$ (Std/ReLU), $87.9\pm0.4$ versus
$87.6\pm0.4$ (Std/GELU), and $87.9\pm0.2$ versus $87.8\pm0.3$ (FFCL/GELU).
The extra score is therefore unnecessary for the central comparison.

\subsection{Goodness Function Comparison}
\label{sec:goodness_comparison}

\textbf{How do the candidate statistics compare in the exploratory grid?}
Table~\ref{tab:goodness} summarizes the single-seed grid, selecting activation
and norm-gate settings per entry. This is descriptive evidence, with untuned SoS and
concatenated scoring, rather than a matched objective comparison.
Burstiness and selective objectives perform well, while the FFCL interaction varies
by goodness. LN-burstiness is consolidated with burstiness in the main comparison;
raw numerical variants are retained only in the appendix for traceability.

\begin{table}[H]
\centering
\caption{Exploratory best test accuracy (\%) per goodness function on Fashion-MNIST (4$\times$2000).
$^\dagger$From \citet{shah2025goodness}.
$^\ddagger$GELU, $\alpha{=}1.5$.
$^\S$Parameter-free. Best activation/norm-gate per entry; seed 42; ensemble scoring.
Std: input-only labels. Burstiness includes the equivalent LN implementation.}
\label{tab:goodness}
\vspace{0.5em}
\small
\begin{tabular}{l cc}
\toprule
\textbf{Goodness function} & \textbf{Standard} & \textbf{FFCL} \\
\midrule
SoS~\citep{hinton2022forward} & 61.4 & 82.4 \\
Variance & 61.5 & 81.7 \\
Game-theoretic$^\dagger$ & 61.4 & 82.4 \\
Neg.\ entropy & 67.4 & 80.4 \\
Softmax-energy-margin$^\dagger$ & 69.8 & 81.9 \\
Contrast top-$k$ & 70.5 & 83.6 \\
Top-$k$ & 79.0 & 82.9 \\
LayerNorm-top-$k$ & 83.3 & 82.8 \\
Entmax-1.5 energy$^\ddagger$ & 85.1 & 87.1 \\
\rowcolor{bestcell}
\textbf{Burstiness (excess kurtosis)$^\S$} & \textbf{88.1} & \textbf{88.4} \\
\bottomrule
\end{tabular}
\end{table}

\subsection{Sweep Analysis: Selectivity, Sparsity, and Moment Order}
\label{sec:sparsity_sweep}

\textbf{How much selectivity or moment sensitivity is useful?}
We sweep three axes on Fashion-MNIST (Figure~\ref{fig:sparsity}):
$k$ (contrast top-$k$), $\alpha$ (entmax), and moment order $p$.
The $k$ sweep has a broad FFCL plateau and weak non-monotonicity under standard FF;
the $\alpha$ and $p$ sweeps have intermediate optima. These are sensitivity studies,
not independent proof of the design hypothesis.

The $k$-sweep (Figure~\ref{fig:sparsity}a) shows FFCL is remarkably robust:
accuracy varies by only 1.7pp (81.9--83.6\%) across a 40$\times$ range.
Standard FF is more sensitive, peaking broadly around $k{=}2$--$5\%$.
The $\alpha$-sweep (Figure~\ref{fig:sparsity}b) shows a clear inverted-U peaking at
$\alpha{\approx}1.5$ (85.1\% Std, 87.1\% FFCL);
at $\alpha{=}1$ (softmax), FFCL diverges entirely (23.6\%), showing failure of this particular dense-weighting configuration.
The moment-$p$ sweep (Figure~\ref{fig:sparsity}c) is the most striking:
at $p{=}2$ (normalized variance), both pathways achieve only $\sim$10\% (chance),
as expected: the normalized second central moment is constant by construction.
This is a sanity check, not evidence that second-order information generally fails.
Performance peaks at $p{=}5$--$6$ (FFCL: \textbf{89.0\%} at $p{=}6$)
and degrades at $p{=}8$; gradient instability is a possible explanation;
FFCL tolerates higher $p$ (84.5\% vs.\ 73.9\% Std), consistent with its
general robustification effect.

\begin{figure}[t]
\centering
\begin{tikzpicture}
\begin{axis}[
    name=kplot,
    width=0.30\columnwidth, height=5cm,
    xlabel={$k$ (\% of $d$)},
    ylabel={Test accuracy (\%)},
    ylabel style={font=\small},
    xlabel style={font=\small},
    xmode=log,
    log basis x=10,
    xtick={0.25, 1, 5, 20},
    xticklabels={0.25, 1, 5, 20},
    x tick label style={font=\scriptsize},
    y tick label style={font=\scriptsize},
    ymin=0, ymax=100,
    grid=major,
    grid style={line width=.1pt, draw=gray!30},
    legend style={at={(0.03,0.97)}, anchor=north west, font=\tiny,
                  draw=gray!50, fill=white, fill opacity=0.9},
    clip=false,
    title={\footnotesize\textbf{(a)} $k$-sweep},
    title style={at={(0.5,1.02)}},
]
\addplot[thick, mark=square*, topkBlue, mark size=1.5pt] coordinates {
    (0.25, 70.97) (0.5, 65.40) (1, 70.49) (2, 71.71)
    (5, 71.77) (10, 68.08) (20, 69.18)
};
\addlegendentry{Standard}
\addplot[thick, mark=*, contrastOrange, dashed, mark size=1.5pt] coordinates {
    (0.25, 83.13) (0.5, 83.42) (1, 83.59) (2, 83.63)
    (5, 83.42) (10, 83.10) (20, 81.89)
};
\addlegendentry{FFCL}
\end{axis}

\begin{axis}[
    at={(kplot.east)}, anchor=west, xshift=0.6cm,
    name=alphaplot,
    width=0.30\columnwidth, height=5cm,
    xlabel={$\alpha$ (entmax)},
    ylabel style={font=\small},
    xlabel style={font=\small},
    xtick={1.0, 1.5, 2.0},
    x tick label style={font=\scriptsize},
    y tick label style={font=\scriptsize},
    ymin=0, ymax=100,
    grid=major,
    grid style={line width=.1pt, draw=gray!30},
    legend style={at={(0.97,0.03)}, anchor=south east, font=\tiny,
                  draw=gray!50, fill=white, fill opacity=0.9},
    clip=false,
    title={\footnotesize\textbf{(b)} $\alpha$-sweep},
    title style={at={(0.5,1.02)}},
]
\addplot[thick, mark=square*, topkBlue, mark size=1.5pt] coordinates {
    (1.0, 74.44) (1.25, 84.87) (1.5, 85.08)
    (1.75, 82.21) (2.0, 78.36)
};
\addlegendentry{Standard}
\addplot[thick, mark=*, contrastOrange, dashed, mark size=1.5pt] coordinates {
    (1.0, 23.60) (1.25, 86.77) (1.5, 87.12)
    (1.75, 85.36) (2.0, 84.41)
};
\addlegendentry{FFCL}
\end{axis}

\begin{axis}[
    at={(alphaplot.east)}, anchor=west, xshift=0.6cm,
    width=0.30\columnwidth, height=5cm,
    xlabel={$p$ (moment order)},
    ylabel style={font=\small},
    xlabel style={font=\small},
    xtick={2, 4, 6, 8},
    x tick label style={font=\scriptsize},
    y tick label style={font=\scriptsize},
    ymin=0, ymax=100,
    grid=major,
    grid style={line width=.1pt, draw=gray!30},
    legend style={at={(0.97,0.03)}, anchor=south east, font=\tiny,
                  draw=gray!50, fill=white, fill opacity=0.9},
    clip=false,
    title={\footnotesize\textbf{(c)} Moment-$p$ sweep},
    title style={at={(0.5,1.02)}},
]
\addplot[thick, mark=square*, topkBlue, mark size=1.5pt] coordinates {
    (2, 10.57) (3, 86.29) (4, 86.98) (5, 87.96) (6, 86.74) (8, 73.89)
};
\addlegendentry{Standard}
\addplot[thick, mark=*, contrastOrange, dashed, mark size=1.5pt] coordinates {
    (2, 10.00) (3, 85.15) (4, 88.41) (5, 88.55) (6, 89.04) (8, 84.54)
};
\addlegendentry{FFCL}
\end{axis}
\end{tikzpicture}
\caption{\textbf{Three sweep axes on Fashion-MNIST (4$\times$2000).}
The $k$ sweep has a broad FFCL plateau; $\alpha$ and $p$ have intermediate optima.
\textbf{(a)}~$k$-sweep: FFCL robust ($<$2pp).
\textbf{(b)}~$\alpha$-sweep: peaks at $\alpha{\approx}1.5$.
\textbf{(c)}~Moment-$p$: peaks at $p{\approx}5$--$6$ (89.04\% FFCL).}
\label{fig:sparsity}
\end{figure}

\citet{shah2025goodness} reported 82.8\% on Fashion-MNIST under a different
pipeline with label forwarding. The held-out selection in
Table~\ref{tab:matched} chose scale 10 and $\tau=1$ for ReLU SoS.
We distinguish published pipeline results from our controlled objective comparisons.

\subsection{Goodness $\times$ Activation Interaction}
\label{sec:interaction}

\textbf{Does the activation change the objective ranking?}
Table~\ref{tab:interaction} reports the single-seed interaction grid.
The rankings vary across activation functions. The matched ReLU and GELU
comparisons in Table~\ref{tab:matched} independently evaluate the objective effect
under equal validation budgets. The mechanisms behind these interactions remain
to be studied.

\begin{table}[H]
\centering
\caption{Exploratory goodness$\times$activation interaction (Fashion-MNIST,
4$\times$2000, standard FF, seed 42, ensemble scoring). Bold: best goodness per row.}
\label{tab:interaction}
\vspace{0.5em}
\small
\begin{tabular}{l cccc}
\toprule
& SoS & Top-$k$ & Entmax-1.5 & Burstiness \\
\midrule
GELU     & 61.4 & 77.7 & 85.1 & \textbf{87.2} \\
Swish    & 56.0 & 79.0 & --- & \textbf{88.1} \\
LN-GELU  & 61.5 & 84.0 & 85.4 & \textbf{87.7} \\
LN-Swish & 65.1 & 84.3 & 85.4 & \textbf{88.0} \\
\bottomrule
\end{tabular}
\end{table}

\subsection{Controlled Loss Comparisons and Spatial Transfer}
\label{sec:spatial}

\textbf{Does the advantage survive alternative losses and spatial blocks?}
In a separate controlled five-seed standard-FF protocol, replacing the loss with
SymBa~\citep{lee2023symba} or SCFF~\citep{chen2025scff} formulations yields
Table~\ref{tab:losses}. These are loss substitutions in our shared setup, not
reproductions of the complete methods, their negative sampling, or their published accuracies.
This protocol is reported separately from the activation/pathway decomposition in
Table~\ref{tab:matched}; its burstiness result is not interchangeable with that table.

\begin{table}[H]
\centering
\caption{Controlled loss/statistic comparison (\%, mean $\pm$ standard deviation,
five seeds; standard per-layer scoring). Method names denote loss substitutions.}
\label{tab:losses}
\small
\begin{tabular}{lcc}
\toprule
Loss/statistic & Fashion-MNIST & MNIST \\
\midrule
FF / tuned SoS & $73.2\pm0.8$ & $87.7\pm0.5$ \\
SymBa loss / energy & $73.6\pm0.2$ & $90.7\pm0.1$ \\
SCFF loss / energy & $73.6\pm0.2$ & $90.7\pm0.1$ \\
FF / burstiness & $87.3\pm0.3$ & $97.5\pm0.1$ \\
\bottomrule
\end{tabular}
\end{table}

For CIFAR-10, both objectives receive the same 18-configuration validation search
on Conv8 with standard labels for 10 epochs. The selected hyperparameters transfer
unchanged to five-seed, 30-epoch Conv8 and block-local ResNet-18 evaluations
(Table~\ref{tab:spatial}). GELU and raw per-block goodness-sum scoring are matched;
outputs are detached at local block boundaries. The double-digit improvements
test architectural transfer, not the convolutional FF accuracy frontier.

\begin{table}[H]
\centering
\caption{CIFAR-10 spatial transfer (\%, mean $\pm$ standard deviation, five seeds).
Std: input-only labels; separate: per-block label pathway.}
\label{tab:spatial}
\small
\begin{tabular}{llccc}
\toprule
Architecture & Labels & SoS & Burstiness & Gain (pp) \\
\midrule
Conv8 & Std & $27.7\pm0.6$ & $40.6\pm1.8$ & +12.9 \\
Conv8 & Separate & $25.0\pm0.7$ & $46.2\pm0.7$ & +21.2 \\
ResNet-18 & Separate & $28.9\pm0.5$ & $46.8\pm2.5$ & +17.8 \\
\bottomrule
\end{tabular}
\end{table}

\subsection{Label-Pathway Interaction}
\label{sec:ffcl_lift}

\textbf{How does label forwarding interact with goodness?}
In the matched GELU comparisons in Table~\ref{tab:matched}, burstiness outperforms
SoS under both standard input-only labels and separate per-layer label forwarding.
The size of the gap depends on the pathway, so the effect of label injection and
the goodness function should be assessed together.

\subsection{Cross-Dataset Generalization}
\label{sec:cross_dataset}

\textbf{How do the objectives behave across datasets?}
Table~\ref{tab:cross_dataset} reports single-seed results across six datasets
under the exploratory ensemble-scoring protocol. The listed configurations differ
in activation and label pathway, and therefore do not isolate goodness effects.
Burstiness configurations achieve strong accuracy across the tested datasets;
matched comparisons in Tables~\ref{tab:matched} and~\ref{tab:spatial} provide
the controlled tests of objective effects.

\begin{table}[H]
\centering
\caption{Exploratory test accuracy (\%) across six datasets (4$\times$2000,
seed~42, ensemble scoring). Rows use different activations and label pathways.}
\label{tab:cross_dataset}
\vspace{0.5em}
\scriptsize
\begin{tabular}{l cccccc}
\toprule
& \textbf{MNIST} & \textbf{F-MNIST} & \textbf{CIFAR-10} & \textbf{USPS} & \textbf{SVHN} & \textbf{EMNIST} \\
\midrule
FFCL + SoS (GELU)   & 93.5 & 82.4 & 46.5 & 76.8 & 71.7 & 77.8 \\
Std + Top-$k$ (Swish) & 90.2 & 78.6 & 37.4 & 82.9 & 57.7 & 66.3 \\
FFCL + Ctrst.\ top-$k$ & 93.3 & 83.6 & 47.1 & 90.0 & 66.0 & 79.7 \\
Std + Burstiness (Swish) & 96.5 & 88.1 & 53.6 & 93.1 & 78.8 & 86.2 \\
FFCL + Burstiness (GELU) & 98.1 & 88.4 & 54.4 & 93.3 & 82.8 & \textbf{91.6} \\
\midrule
Std + Burstiness (LN-GELU) & 96.9 & 87.7 & 53.3 & \textbf{93.9} & 81.7 & 89.0 \\
FFCL + Burstiness (LN-GELU) & \textbf{98.2} & \textbf{88.7} & 54.8 & 93.4 & \textbf{84.0} & 91.2 \\
Std + Burstiness (LN-Swish) & 97.2 & 88.0 & 54.3 & 92.7 & 82.1 & 89.1 \\
\rowcolor{bestcell}
FFCL + Burstiness (LN-Swish) & 98.1 & 88.7 & \textbf{55.6} & 92.6 & 83.7 & 91.1 \\
\bottomrule
\end{tabular}
\end{table}

\section{Analysis}
\label{sec:analysis}

Our matched results support a scoped design hypothesis: activity-shape statistics
can provide useful local learning signals beyond energy in the tested FF settings.
This is an existence result, not a necessary or sufficient condition for effective
goodness, nor a universal law across architectures and datasets. Selective statistics
still retain magnitude information, whereas kurtosis removes uniform scale.

The ICA connection is motivational: kurtosis is a classical non-Gaussianity criterion
in ICA~\citep{hyvarinen2000independent}, but maximizing coordinate-wise kurtosis in
supervised FF does not establish independent components or perform a full ICA algorithm.
Likewise, intermediate sweep optima can arise from ordinary hyperparameter sensitivity.
The constant $p=2$ statistic is only a sanity check.
A falsifiable future diagnostic is an orthogonal rotation of the activation coordinates:
it preserves SoS energy while generally changing coordinate-wise kurtosis. We have not
run that diagnostic or a dedicated shape-sensitive negative control.

\paragraph{Activation interaction.}
The exploratory grid shows that the preferred activation depends on the goodness
statistic; the tuned comparison also finds a weaker SoS result with GELU than ReLU.
ReLU sparsity and smooth-activation density offer a possible interpretation, but we
do not directly measure the proposed signal-dilution mechanism. In particular,
shape-sensitive objectives are not guaranteed to eliminate background noise.


\section{Discussion and Limitations}
\label{sec:limitations}

\paragraph{Accuracy and scope.}
Matched goodness-only comparisons provide the primary evidence. Published
backpropagation and FF accuracies use different pipelines and cannot be directly
compared without matching their training and evaluation protocols. The six-dataset grid remains exploratory and
mostly single-seed; only subsets receive repeated evaluation. Spatial transfer is
limited to CIFAR-10 local Conv8 and ResNet-18; larger-scale generalization remains open.

\paragraph{Reproducibility and cost.}
The matched MLP and spatial tests use five seeds; the broad combinatorial grid uses
seed 42 and concatenated-ensemble scoring. Standard per-layer scoring is primary.
Candidate-label inference still requires a forward pass for each class, limiting
scalability as class count grows. In the spatial experiments burstiness increases
inference time by approximately 16--21\% and training time by 24--29\% over SoS
under the same candidate-label protocol; these are setting-specific costs.
The statistic has no shape hyperparameter, but the training protocol still requires tuning.

\paragraph{Hyperparameters.}
Top-$k$ and entmax each introduce one hyperparameter, but our sweeps show
FFCL configurations are remarkably robust ($<$2pp variation for $k$, $<$3pp
for $\alpha$).
Burstiness is entirely parameter-free.
The moment-$p$ family has $p{\in}[4,6]$ as a broad, stable optimum.

\section{Conclusion}
\label{sec:conclusion}

The choice of goodness function substantially affects Forward-Forward learning.
We studied selective objectives that focus on peak activities and shape-sensitive
objectives based on normalized moments, including burstiness, a parameter-free and
scale-invariant excess-kurtosis statistic. Our experiments show that these choices
interact with activation functions and label pathways, motivating comparisons that
hold these components fixed when assessing an objective.

In validation-tuned, five-seed Fashion-MNIST experiments, burstiness improves
accuracy from $73.2\pm0.9\%$ to $88.6\pm0.4\%$ over SoS with matched ReLU
activations, standard label injection, and per-layer scoring. The advantage also
persists in controlled comparisons with separate per-layer labels and in local
convolutional and residual blocks on CIFAR-10. Together, these findings show
that activity-shape statistics can provide effective local learning signals across
the tested settings and architectures.

The broader goodness-function grid and parameter sweeps reveal useful variation
across objectives, while also highlighting sensitivity to experimental choices.
Future work can test shape-sensitive objectives on larger datasets and architectures,
examine their behavior under changes of activation coordinates, and study their
interaction with alternative local learning procedures and inference strategies.

\section*{Acknowledgements and Disclosures}
The authors received no third-party funding or support for the research reported
in this paper. We thank Poindexter Labs for its support for in-person attendance at
the conference. The authors declare no other competing interests.

\bibliographystyle{plainnat}
\bibliography{references}

@misc{hinton2022forward,
      title={The Forward-Forward Algorithm: Some Preliminary Investigations}, 
      author={Geoffrey Hinton},
      year={2022},
      eprint={2212.13345},
      archivePrefix={arXiv},
      primaryClass={cs.LG},
      url={https://arxiv.org/abs/2212.13345}, 
}

@INPROCEEDINGS{10191727,
  author={Giampaolo, Fabio and Izzo, Stefano and Prezioso, Edoardo and Piccialli, Francesco},
  booktitle={2023 International Joint Conference on Neural Networks (IJCNN)}, 
  title={Investigating Random Variations of the Forward-Forward Algorithm for Training Neural Networks}, 
  year={2023},
  volume={},
  number={},
  pages={1-7},
  doi={10.1109/IJCNN54540.2023.10191727}}

@misc{lorberbom2024layer,
      title={Layer Collaboration in the Forward-Forward Algorithm}, 
      author={Guy Lorberbom and Itai Gat and Yossi Adi and Alex Schwing and Tamir Hazan},
      year={2023},
      eprint={2305.12393},
      archivePrefix={arXiv},
      primaryClass={cs.LG},
      url={https://arxiv.org/abs/2305.12393}, 
}

@misc{lee2023symba,
      title={SymBa: Symmetric Backpropagation-Free Contrastive Learning with Forward-Forward Algorithm for Optimizing Convergence}, 
      author={Heung-Chang Lee and Jeonggeun Song},
      year={2023},
      eprint={2303.08418},
      archivePrefix={arXiv},
      primaryClass={cs.CV},
      url={https://arxiv.org/abs/2303.08418}, 
}

@misc{ororbia2023predictive,
      title={The Predictive Forward-Forward Algorithm}, 
      author={Alexander Ororbia and Ankur Mali},
      year={2023},
      eprint={2301.01452},
      archivePrefix={arXiv},
      primaryClass={cs.LG},
      url={https://arxiv.org/abs/2301.01452}, 
}

@misc{hendrycks2016gelu,
      title={Gaussian Error Linear Units (GELUs)}, 
      author={Dan Hendrycks and Kevin Gimpel},
      year={2023},
      eprint={1606.08415},
      archivePrefix={arXiv},
      primaryClass={cs.LG},
      url={https://arxiv.org/abs/1606.08415}, 
}

@misc{ramachandran2017searching,
      title={Searching for Activation Functions}, 
      author={Prajit Ramachandran and Barret Zoph and Quoc V. Le},
      year={2017},
      eprint={1710.05941},
      archivePrefix={arXiv},
      primaryClass={cs.NE},
      url={https://arxiv.org/abs/1710.05941}, 
}

@article{olshausen1996emergence,
  author   = {Olshausen, Bruno A. and Field, David J.},
  title    = {Emergence of simple-cell receptive field properties by learning a sparse code for natural images},
  journal  = {Nature},
  volume   = {381},
  number   = {6583},
  pages    = {607--609},
  year     = {1996},
  month    = {jun},
  issn     = {1476-4687},
  doi      = {10.1038/381607a0},
  url      = {https://doi.org/10.1038/381607a0}
}

@ARTICLE{lecun1998mnist,
  author={Lecun, Y. and Bottou, L. and Bengio, Y. and Haffner, P.},
  journal={Proceedings of the IEEE}, 
  title={Gradient-based learning applied to document recognition}, 
  year={1998},
  volume={86},
  number={11},
  pages={2278-2324},
  doi={10.1109/5.726791}}

@misc{xiao2017fashionmnist,
      title={Fashion-MNIST: a Novel Image Dataset for Benchmarking Machine Learning Algorithms}, 
      author={Han Xiao and Kashif Rasul and Roland Vollgraf},
      year={2017},
      eprint={1708.07747},
      archivePrefix={arXiv},
      primaryClass={cs.LG},
      url={https://arxiv.org/abs/1708.07747}, 
}

@misc{kingma2015adam,
      title={Adam: A Method for Stochastic Optimization}, 
      author={Diederik P. Kingma and Jimmy Ba},
      year={2017},
      eprint={1412.6980},
      archivePrefix={arXiv},
      primaryClass={cs.LG},
      url={https://arxiv.org/abs/1412.6980}, 
}

@Book{hebb1949organization,
author={Hebb, D. O.},
title={The organization of behavior; a neuropsychological theory.},
series={The organization of behavior; a neuropsychological theory.},
year={1949},
publisher={Wiley},
address={Oxford,  England},
pages={xix, 335-xix, 335}
}

@article{xie2003equivalence,
  author   = {Xie, Xiaohui and Seung, H. Sebastian},
  title    = {Equivalence of backpropagation and contrastive Hebbian learning in a layered network},
  journal  = {Neural Computation},
  volume   = {15},
  number   = {2},
  pages    = {441--454},
  year     = {2003},
  month    = {feb},
  doi      = {10.1162/089976603762552988}
}

@misc{scellier2017equilibrium,
      title={Equilibrium Propagation: Bridging the Gap Between Energy-Based Models and Backpropagation}, 
      author={Benjamin Scellier and Yoshua Bengio},
      year={2017},
      eprint={1602.05179},
      archivePrefix={arXiv},
      primaryClass={cs.LG},
      url={https://arxiv.org/abs/1602.05179}, 
}

@misc{ahmad2019dense,
      title={How Can We Be So Dense? The Benefits of Using Highly Sparse Representations}, 
      author={Subutai Ahmad and Luiz Scheinkman},
      year={2019},
      eprint={1903.11257},
      archivePrefix={arXiv},
      primaryClass={cs.LG},
      url={https://arxiv.org/abs/1903.11257}, 
}

@article{maass2000computational,
  author   = {Maass, W.},
  title    = {On the computational power of winner-take-all},
  journal  = {Neural Computation},
  volume   = {12},
  number   = {11},
  pages    = {2519--2535},
  year     = {2000},
  month    = {nov},
  doi      = {10.1162/089976600300014827}
}

@misc{shah2025goodness,
      title={In Search of Goodness: Large Scale Benchmarking of Goodness Functions for the Forward-Forward Algorithm}, 
      author={Arya Shah and Vaibhav Tripathi},
      year={2025},
      eprint={2511.18567},
      archivePrefix={arXiv},
      primaryClass={cs.LG},
      url={https://arxiv.org/abs/2511.18567}, 
}

@misc{srinivasan2024forward,
      title={FFCL: Forward-Forward Net with Cortical Loops, Training and Inference on Edge Without Backpropagation}, 
      author={Ali Karkehabadi and Houman Homayoun and Avesta Sasan},
      year={2024},
      eprint={2405.12443},
      archivePrefix={arXiv},
      primaryClass={cs.LG},
      url={https://arxiv.org/abs/2405.12443}, 
}

@misc{correia2019adaptively,
      title={Adaptively Sparse Transformers}, 
      author={Gonçalo M. Correia and Vlad Niculae and André F. T. Martins},
      year={2019},
      eprint={1909.00015},
      archivePrefix={arXiv},
      primaryClass={cs.CL},
      url={https://arxiv.org/abs/1909.00015}, 
}

@misc{peters2019sparse,
      title={Sparse Sequence-to-Sequence Models}, 
      author={Ben Peters and Vlad Niculae and André F. T. Martins},
      year={2019},
      eprint={1905.05702},
      archivePrefix={arXiv},
      primaryClass={cs.CL},
      url={https://arxiv.org/abs/1905.05702}, 
}

@misc{martins2016softmax,
      title={From Softmax to Sparsemax: A Sparse Model of Attention and Multi-Label Classification}, 
      author={André F. T. Martins and Ramón Fernandez Astudillo},
      year={2016},
      eprint={1602.02068},
      archivePrefix={arXiv},
      primaryClass={cs.CL},
      url={https://arxiv.org/abs/1602.02068}, 
}

@article{hyvarinen2000independent,
title = {Independent component analysis: algorithms and applications},
journal = {Neural Networks},
volume = {13},
number = {4},
pages = {411-430},
year = {2000},
issn = {0893-6080},
doi = {https://doi.org/10.1016/S0893-6080(00)00026-5},
url = {https://www.sciencedirect.com/science/article/pii/S0893608000000265},
author = {A. Hyvärinen and E. Oja}
}

@article{lisman1997bursts,
  author   = {Lisman, J. E.},
  title    = {Bursts as a unit of neural information: making unreliable synapses reliable},
  journal  = {Trends in Neurosciences},
  volume   = {20},
  number   = {1},
  pages    = {38--43},
  year     = {1997},
  month    = {jan},
  doi      = {10.1016/S0166-2236(96)10070-9}
}

@misc{martin2019traditional,
      title={Traditional and Heavy-Tailed Self Regularization in Neural Network Models}, 
      author={Charles H. Martin and Michael W. Mahoney},
      year={2019},
      eprint={1901.08276},
      archivePrefix={arXiv},
      primaryClass={cs.LG},
      url={https://arxiv.org/abs/1901.08276}, 
}

@article{krizhevsky2009learning,
author = {Krizhevsky, Alex},
year = {2012},
month = {05},
pages = {},
title = {Learning Multiple Layers of Features from Tiny Images},
journal = {University of Toronto}
}

@ARTICLE{hull1994database,
  author={Hull, J.J.},
  journal={IEEE Transactions on Pattern Analysis and Machine Intelligence}, 
  title={A database for handwritten text recognition research}, 
  year={1994},
  volume={16},
  number={5},
  pages={550-554},
  doi={10.1109/34.291440}}

@article{netzer2011reading,
author = {Netzer, Yuval and Wang, Tao and Coates, Adam and Bissacco, Alessandro and Wu, Bo and Ng, Andrew},
year = {2011},
month = {01},
pages = {},
title = {Reading Digits in Natural Images with Unsupervised Feature Learning},
journal = {NIPS}
}

@INPROCEEDINGS{cohen2017emnist,
  author={Cohen, Gregory and Afshar, Saeed and Tapson, Jonathan and van Schaik, André},
  booktitle={2017 International Joint Conference on Neural Networks (IJCNN)}, 
  title={EMNIST: Extending MNIST to handwritten letters}, 
  year={2017},
  volume={},
  number={},
  pages={2921-2926},
  doi={10.1109/IJCNN.2017.7966217}}

@Article{chen2025scff,
author={Chen, Xing
and Liu, Dongshu
and Laydevant, J{\'e}r{\'e}mie
and Grollier, Julie},
title={Self-Contrastive Forward-Forward algorithm},
journal={Nature Communications},
year={2025},
month={Jul},
day={01},
volume={16},
number={1},
pages={5978},
issn={2041-1723},
doi={10.1038/s41467-025-61037-0},
url={https://doi.org/10.1038/s41467-025-61037-0}
}

@misc{gong2025mono,
      title={Mono-Forward: Revisiting Forward-Forward through Objective-Locality Decomposition}, 
      author={James Gong and Bruce Li and Waleed Abdulla},
      year={2026},
      eprint={2501.09238},
      archivePrefix={arXiv},
      primaryClass={cs.LG},
      url={https://arxiv.org/abs/2501.09238}, 
}

@article{scodellaro2023conv,
   title={Training convolutional neural networks with the Forward–Forward Algorithm},
   volume={15},
   ISSN={2045-2322},
   url={http://dx.doi.org/10.1038/s41598-025-26235-2},
   DOI={10.1038/s41598-025-26235-2},
   number={1},
   journal={Scientific Reports},
   publisher={Springer Science and Business Media LLC},
   author={Scodellaro, Riccardo and Kulkarni, Ajinkya and Alves, Frauke and Schröter, Matthias},
   year={2025},
   month=Nov }

@misc{krutsylo2025scalable,
      title={Scalable Forward-Forward Algorithm}, 
      author={Andrii Krutsylo},
      year={2025},
      eprint={2501.03176},
      archivePrefix={arXiv},
      primaryClass={cs.LG},
      url={https://arxiv.org/abs/2501.03176}, 
}

@misc{gong2025asge,
      title={Adaptive Spatial Goodness Encoding: Advancing and Scaling Forward-Forward Learning Without Backpropagation}, 
      author={Qingchun Gong and Robert Bogdan Staszewski and Kai Xu},
      year={2026},
      eprint={2509.12394},
      archivePrefix={arXiv},
      primaryClass={cs.LG},
      url={https://arxiv.org/abs/2509.12394}, 
}

\newpage
\appendix

\section{Experimental Details}
\label{app:details}

\paragraph{Label embedding.}
Following \citet{hinton2022forward}, input images are flattened to a 784-dimensional vector
and concatenated with a one-hot label vector scaled by $s = 5.0$, yielding a 794-dimensional
input that is L2-normalized.
For FFCL, the input to the first layer uses only the 784-dimensional image (no label
concatenation), and labels are injected at every layer via the per-layer label projection
(Eq.~\ref{eq:ffcl}).

\paragraph{Top-$k$ parameter.}
For the 4$\times$2000 architecture, $k = \max(5, \lfloor 0.02 \times 2000\rfloor) = 40$
neurons (2\% of layer width).
For contrast top-$k$, $k = \max(5, \lfloor 0.01 \times 2000\rfloor) = 20$ neurons (1\%).
The sparsity sweep in \S\ref{sec:sparsity_sweep} varies $k$ from 0.25\% to 20\%
(5 to 400 neurons).

\paragraph{Norm-gating.}
Norm-gating applies $\sigma(\|\mathbf{h}\|) \cdot \mathbf{h}$ after activation, where
$\sigma$ is the sigmoid function.
This rescales the activation vector based on its norm before passing to the next layer.
In all experiments, the maximum accuracy difference between norm-gate on/off
(same goodness/activation) was 0.39pp; we report the best of each pair in the main text.

\paragraph{Entmax computation.}
We use the \texttt{entmax} package~\citep{correia2019adaptively} which implements
$\alpha$-entmax via the bisection algorithm.
For the sparsity sweep, $\alpha$ ranges from 1.0 (softmax) to 2.0 (sparsemax) in steps
of 0.25.
Entmax-weighted energy (\S\ref{sec:entmax}) first applies $\alpha$-entmax to the activation
vector to obtain sparse weights, then computes the weighted sum of squared activations.

\paragraph{External baseline implementations.}
\emph{Softmax-energy-margin}~\citep{shah2025goodness}: We implement this as
$g = g_{\text{SoS}} + 0.5 \cdot \bar{g}_{\text{SoS}} \cdot
(-\log\sum_i\exp(h_i / T))$ with temperature $T = 1.0$ and margin $\lambda = 0.5$,
where $\bar{g}_{\text{SoS}}$ is a running mean of goodness values.
\emph{Game-theoretic}: We implement this as a weighted SoS where each neuron's squared
activation is weighted by its relative magnitude
$w_i = |h_i| / (\sum_j |h_j| + \epsilon)$.

\paragraph{Compute resources.}
Experiments were run on NVIDIA A100 GPUs.
Each standard FF experiment takes 30--60 seconds.
Each FFCL experiment takes 35--70 seconds (the label projection adds minimal overhead).
Each entmax experiment takes 200--400 seconds due to the bisection-based entmax computation.
The exploratory suite (including all goodness functions, both label pathways,
the sparsity sweep, and both datasets) completes in approximately 4 GPU-hours.

\clearpage
\section{MNIST Results}
\label{app:mnist}

Table~\ref{tab:mnist_full} presents the complete MNIST results (4$\times$2000).
On this task, the best configurations exceed 88\% accuracy, and the differences between
goodness functions are smaller than on Fashion-MNIST.
FFCL provides a consistent ${\sim}$3--5pp lift.

\begin{table}[H]
\centering
\caption{Test accuracy (\%) on MNIST (4$\times$2000), ranked by accuracy.
Best of norm-gate on/off reported for each configuration.
\colorbox{bestcell}{Highlighted}: burstiness variants.}
\label{tab:mnist_full}
\vspace{0.5em}
\scriptsize
\begin{tabular}{r l l l r}
\toprule
\# & Label & Act. & Goodness & Acc\% \\
\midrule
\rowcolor{bestcell} 1 & FFCL & GELU & burstiness & \textbf{98.09} \\
\rowcolor{bestcell} 2 & FFCL & GELU & LN-burstiness & 98.09 \\
\rowcolor{bestcell} 3 & FFCL & Swish & burstiness & 98.00 \\
\rowcolor{bestcell} 4 & FFCL & Swish & LN-burstiness & 98.00 \\
\rowcolor{bestcell} 5 & Std & Swish & burstiness & 96.54 \\
\rowcolor{bestcell} 6 & Std & Swish & LN-burstiness & 96.45 \\
7 & Std & GELU & LN-burstiness & 93.64 \\
8 & FFCL & GELU & SoS & 93.45 \\
9 & FFCL & GELU & game-theoretic & 93.45 \\
10 & FFCL & GELU & contrast top-$k$ & 93.34 \\
11 & FFCL & GELU & variance & 93.22 \\
12 & FFCL & Swish & game-theoretic & 93.22 \\
13 & FFCL & Swish & SoS & 93.21 \\
14 & FFCL & Swish & contrast top-$k$ & 93.14 \\
15 & FFCL & GELU & top-$k$ & 92.99 \\
16 & FFCL & Swish & top-$k$ & 92.93 \\
17 & Std & GELU & burstiness & 92.49 \\
18 & FFCL & Swish & variance & 92.23 \\
19 & Std & GELU & LN-top-$k$ & 92.06 \\
20 & FFCL & GELU & softmax-e-m & 91.61 \\
21 & Std & Swish & LN-top-$k$ & 90.80 \\
22 & FFCL & Swish & softmax-e-m & 90.40 \\
23 & Std & GELU & top-$k$ & 90.37 \\
24 & Std & Swish & top-$k$ & 90.23 \\
25 & FFCL & GELU & LN-top-$k$ & 89.34 \\
26 & FFCL & Swish & LN-top-$k$ & 89.08 \\
27 & Std & ReLU & SoS (baseline) & 88.76 \\
28 & FFCL & Swish & entropy & 88.50 \\
29 & FFCL & GELU & entropy & 88.17 \\
30 & Std & GELU & contrast top-$k$ & 86.90 \\
31 & Std & Swish & contrast top-$k$ & 86.45 \\
32 & Std & Swish & SoS & 82.54 \\
33 & Std & Swish & game-theoretic & 82.51 \\
34 & Std & GELU & game-theoretic & 78.06 \\
35 & Std & GELU & SoS & 77.93 \\
\bottomrule
\end{tabular}
\end{table}

In this single-seed exploratory grid, FFCL + GELU + burstiness achieves 98.09\%
on MNIST. The validation-tuned, five-seed MNIST comparison is reported in
Table~\ref{tab:losses}.

\clearpage
\section{Complete Fashion-MNIST Results}
\label{app:fashion_full}

Table~\ref{tab:fashion_full} presents the complete ranked results for all configurations on
Fashion-MNIST (4$\times$2000), including both standard and FFCL label pathways.

\begin{table}[H]
\centering
\caption{Complete ranked results on Fashion-MNIST (4$\times$2000).
Best of norm-gate on/off reported.
\colorbox{bestcell}{Highlighted}: burstiness variants.}
\label{tab:fashion_full}
\vspace{0.5em}
\scriptsize
\begin{tabular}{r l l l r}
\toprule
\# & Label & Act. & Goodness & Acc\% \\
\midrule
\rowcolor{bestcell} 1 & FFCL & GELU & burstiness & \textbf{88.41} \\
\rowcolor{bestcell} 2 & FFCL & GELU & LN-burstiness & 88.41 \\
\rowcolor{bestcell} 3 & FFCL & Swish & burstiness & 88.21 \\
\rowcolor{bestcell} 4 & FFCL & Swish & LN-burstiness & 88.21 \\
\rowcolor{bestcell} 5 & Std & Swish & LN-burstiness & 88.11 \\
\rowcolor{bestcell} 6 & Std & Swish & burstiness & 88.07 \\
\rowcolor{bestcell} 7 & Std & GELU & burstiness & 87.15 \\
\rowcolor{bestcell} 8 & Std & GELU & LN-burstiness & 87.14 \\
9 & FFCL & GELU & contrast top-$k$ & 83.59 \\
10 & FFCL & Swish & contrast top-$k$ & 83.55 \\
11 & Std & GELU & LN-top-$k$ & 83.28 \\
12 & FFCL & Swish & top-$k$ & 82.93 \\
13 & FFCL & Swish & LN-top-$k$ & 82.75 \\
14 & Std & Swish & LN-top-$k$ & 82.71 \\
15 & FFCL & GELU & top-$k$ & 82.65 \\
16 & FFCL & GELU & game-theoretic & 82.38 \\
17 & FFCL & GELU & SoS & 82.38 \\
18 & FFCL & GELU & LN-top-$k$ & 82.25 \\
19 & FFCL & Swish & game-theoretic & 82.18 \\
20 & FFCL & Swish & SoS & 82.17 \\
21 & FFCL & GELU & softmax-e-m & 81.89 \\
22 & FFCL & GELU & variance & 81.74 \\
23 & FFCL & Swish & variance & 81.29 \\
24 & FFCL & Swish & softmax-e-m & 80.98 \\
25 & FFCL & Swish & entropy & 80.43 \\
26 & FFCL & GELU & entropy & 79.93 \\
27 & Std & Swish & top-$k$ & 79.03 \\
28 & Std & GELU & top-$k$ & 77.65 \\
29 & Std & GELU & contrast top-$k$ & 70.49 \\
30 & Std & Swish & softmax-e-m & 69.85 \\
31 & Std & GELU & softmax-e-m & 68.72 \\
32 & Std & Swish & contrast top-$k$ & 67.99 \\
33 & Std & Swish & entropy & 67.39 \\
34 & Std & GELU & entropy & 66.07 \\
35 & Std & GELU & variance & 61.55 \\
36 & Std & GELU & SoS & 61.43 \\
37 & Std & GELU & game-theoretic & 61.37 \\
38 & Std & Swish & variance & 60.72 \\
40 & Std & Swish & SoS & 55.99 \\
41 & Std & Swish & game-theoretic & 55.99 \\
\bottomrule
\end{tabular}
\end{table}

Burstiness and LN-burstiness occupy all top-8 positions. This table does \emph{not}
include entmax or moment-$p$ results, which are presented in the sweep
(Table~\ref{tab:sweep_full}).
When sweep results are included, FFCL + moment-$p\!=\!6$ (89.04\%) takes the overall
top position, with FFCL + moment-$p\!=\!5$ (88.55\%) and FFCL + burstiness (88.41\%)
closely following.

\clearpage
\section{Sparsity Sweep: Full Results}
\label{app:sweep}

Table~\ref{tab:sweep_full} presents the complete numerical results from the sparsity sweep
(Figure~\ref{fig:sparsity}).

\begin{table}[H]
\centering
\caption{Complete sweep results on Fashion-MNIST (4$\times$2000, GELU, no norm-gate).}
\label{tab:sweep_full}
\vspace{0.5em}

\begin{minipage}[t]{0.32\textwidth}
\centering
\small
\textbf{(a) $k$-sweep}\\[3pt]
\scriptsize
\begin{tabular}{r r r}
\toprule
$k$ (\%) & Std & FFCL \\
\midrule
0.25 & 70.97 & 83.13 \\
0.5 & 65.40 & 83.42 \\
1 & 70.49 & 83.59 \\
\textbf{2} & \textbf{71.71} & \textbf{83.63} \\
5 & 71.77 & 83.42 \\
10 & 68.08 & 83.10 \\
20 & 69.18 & 81.89 \\
\midrule
Range & 6.37 & 1.74 \\
\bottomrule
\end{tabular}
\end{minipage}%
\hfill
\begin{minipage}[t]{0.32\textwidth}
\centering
\small
\textbf{(b) $\alpha$-sweep}\\[3pt]
\scriptsize
\begin{tabular}{r r r}
\toprule
$\alpha$ & Std & FFCL \\
\midrule
1.00 & 74.44 & 23.60$^*$ \\
1.25 & 84.87 & 86.77 \\
\textbf{1.50} & \textbf{85.08} & \textbf{87.12} \\
1.75 & 82.21 & 85.36 \\
2.00 & 78.36 & 84.41 \\
\midrule
Range$^\dagger$ & 6.72 & 2.71 \\
\bottomrule
\end{tabular}
\\[2pt]
{\tiny $^*$Diverged. $^\dagger$Excl.\ $\alpha\!=\!1$.}
\end{minipage}%
\hfill
\begin{minipage}[t]{0.32\textwidth}
\centering
\small
\textbf{(c) Moment-$p$ sweep}\\[3pt]
\scriptsize
\begin{tabular}{r r r}
\toprule
$p$ & Std & FFCL \\
\midrule
2 & 10.57$^*$ & 10.00$^*$ \\
3 & 86.29 & 85.15 \\
4 & 86.98 & 88.41 \\
5 & 87.96 & 88.55 \\
\textbf{6} & 86.74 & \textbf{89.04} \\
8 & 73.89 & 84.54 \\
\midrule
Range$^\dagger$ & 14.07 & 4.50 \\
\bottomrule
\end{tabular}
\\[2pt]
{\tiny $^*$Constant-statistic sanity check. $^\dagger$Excl.\ $p\!=\!2$.}
\end{minipage}
\end{table}

\paragraph{Key observations.}
(1)~FFCL is dramatically more robust than standard FF across all three sweep axes:
FFCL ranges are 1.74pp ($k$), 2.71pp ($\alpha$), and 4.50pp ($p$, excl.\ $p\!=\!2$),
compared to 6.37pp, 6.72pp, and 14.07pp for standard.
(2)~The $k$ sweep has a broad FFCL plateau; $\alpha$ and $p$ have intermediate optima.
(3)~The moment-$p$ sweep reveals the highest absolute accuracies (89.04\% at $p\!=\!6$
for FFCL), showing the strongest result within this exploratory sweep.
(4)~At $p=2$ the normalized statistic is constant by construction, so chance
accuracy is a sanity check.

\clearpage
\section{Burstiness: Detailed Results}
\label{app:burstiness}

Table~\ref{tab:burstiness_detail} presents a detailed breakdown of burstiness results
across all activation and label-pathway combinations on Fashion-MNIST (4$\times$2000).

\begin{table}[H]
\centering
\caption{Burstiness variants on Fashion-MNIST (4$\times$2000, best of norm-gate on/off).
LN-burstiness is an equivalent numerical implementation, not a distinct objective.
Raw runs are retained here for traceability. LN-GELU and LN-Swish activations improve over GELU/Swish with FFCL.}
\label{tab:burstiness_detail}
\vspace{0.5em}
\small
\begin{tabular}{l l l r}
\toprule
\textbf{Label} & \textbf{Act.} & \textbf{Goodness} & \textbf{Acc\%} \\
\midrule
\rowcolor{bestcell}
FFCL & LN-GELU & LN-burstiness & \textbf{88.71} \\
FFCL & LN-GELU & burstiness & 88.69 \\
FFCL & LN-Swish & burstiness & 88.68 \\
FFCL & LN-Swish & LN-burstiness & 88.66 \\
FFCL & GELU & burstiness & 88.41 \\
FFCL & GELU & LN-burstiness & 88.41 \\
FFCL & Swish & burstiness & 88.21 \\
FFCL & Swish & LN-burstiness & 88.21 \\
Std & Swish & LN-burstiness & 88.11 \\
Std & Swish & burstiness & 88.07 \\
Std & LN-Swish & burstiness & 87.98 \\
Std & LN-Swish & LN-burstiness & 87.98 \\
Std & LN-GELU & LN-burstiness & 87.67 \\
Std & LN-GELU & burstiness & 87.66 \\
Std & GELU & burstiness & 87.15 \\
Std & GELU & LN-burstiness & 87.14 \\
\bottomrule
\end{tabular}
\end{table}

\paragraph{Key observations.}
(1)~LayerNorm before kurtosis is mathematically redundant for nonzero variance;
small numerical differences in these runs are not evidence for a separate method.
(2)~Pre-activation normalization (LN-GELU, LN-Swish) consistently outperforms plain
GELU/Swish with FFCL: LN-GELU reaches 88.71\% and LN-Swish 88.68\%, vs.\
88.41\% for GELU and 88.21\% for Swish.
For standard FF, LN-Swish (87.98\%) slightly trails Swish (88.07\%) but beats
LN-GELU (87.66\%) and GELU (87.15\%).
(3)~The label-pathway interaction varies by activation in this single-seed grid.


\end{document}